\documentclass[sn-mathphys,Numbered]{sn-jnl}% Math and Physical Sciences Reference Style
\usepackage{graphicx}%
\usepackage{multirow}%
\usepackage{amsmath,amssymb,amsfonts}%
\usepackage{amsthm}%
\usepackage{mathrsfs}%
\usepackage[title]{appendix}%
\usepackage{xcolor}%
\usepackage{textcomp}%
\usepackage{manyfoot}%
\usepackage{booktabs}%
\usepackage{algorithm}%
\usepackage{algorithmicx}%
\usepackage{algpseudocode}%
\usepackage{listings}%
\theoremstyle{thmstyleone}%
\theoremstyle{thmstyletwo}%

\theoremstyle{thmstylethree}%

\begin{document}

\title[Article Title]{Investigating YOLO Models Towards Outdoor Obstacle Detection For Visually Impaired People}

\author[1]{\fnm{Chenhao} \sur{He}}\email{2240318@ncpachina.org}
\equalcont{The authors contributed equally to this work.}

\author*[2]{\fnm{Pramit} \sur{Saha}}\email{pramit.saha@eng.ox.ac.uk}
\equalcont{The authors contributed equally to this work.}

\affil[1]{\orgname{Nansha College Preparatory Academy}, \orgaddress{ \country{China}}}

\affil*[2]{\orgdiv{Department of Engineering Science}, \orgname{University of Oxford}, \orgaddress{ \country{United Kingdom}}}


\abstract{The utilization of deep learning-based object detection is an effective approach to assist visually impaired individuals in avoiding obstacles. In this paper, we implemented seven different YOLO object detection models \textit{viz}., YOLO-NAS (small, medium, large), YOLOv8, YOLOv7, YOLOv6, and YOLOv5 and performed comprehensive evaluation with carefully tuned hyperparameters, to analyze how these models performed on images containing common daily-life objects presented on roads and sidewalks. After a systematic investigation, YOLOv8 was found to be the best model, which reached a precision of $80\%$ and a recall of $68.2\%$ on a well-known Obstacle Dataset which includes images from VOC dataset, COCO dataset, and TT100K dataset along with images collected by the researchers in the field. Despite being the latest model and demonstrating better performance in many other applications, YOLO-NAS was found to be suboptimal for the obstacle detection task.}



\maketitle
\section{Introduction}
Vision is one of the most important senses of our human body. It helps us to identify our surroundings, allowing us to carry out daily work. However, with a loss of vision, people struggle with the basic skills in life, such as the ability to recognize obstacles, learn, read, participate in school, and work. According to the World Health Organization (WHO), at least 2.2 billion people around the world have a near or distant vision impairment\cite{who2023blindness}. Some of the leading causes of blindness and visual impairment are cataracts, glaucoma, undercorrected refractive error, age-related macular degeneration, and diabetic retinopathy \cite{steinmetz2021causes}. There are serious consequences along with visual impairment for individuals, including lower rates of workforce participation and productivity \cite{frick2015global}, higher rates of depression and anxiety \cite{evans2007depression}, and higher rates of experiencing violence and abuse, including bullying and sexual violence \cite{brunes2018bullying}. The economy is also hugely affected, as studies showed that the annual cost of moderate to severe vision impairment ranged from 0.1 billion US dollars in Honduras to as high as 16.5 billion US dollars in the United States of America \cite{eckert2015simple}.

One way to effectively mitigate and resolve this worldwide issue is through the utilization of Deep Learning methods, specifically, through the use of Convolutional Neural Networks (CNN) to guide the visually impaired in carrying out various daily object detection tasks. Visually impaired and blind people can be assisted by CNN because of its algorithmic ability to recognize obstacles in front. The object detection algorithms are able to inform the person of what objects are in front of their way and thus avoid them \cite{tapu2014real}. In the light of this, this work evaluated and analyzed the performance of YOLO models on a popular obstacle dataset containing obstacles and objects on daily streets and sidewalks.

The different YOLO models we have used are YOLO v5, v6, v7, v8, and NAS. NAS refers to Neural Architecture Search, where a neural network automates the process of finding the best architecture for doing the task, rather than doing it by ourselves \cite{elsken2019neural,said2023obstacle}. We have done comprehensive testing and tweaking of the models.

Our work aims to specifically look at how these different versions of YOLO models vary in performance on the obstacle detection task, and also how the newest architecture search algorithm (YOLO-NAS) is compared to the earlier versions. We particularly choose YOLO for this systematic empirical investigation owing to its well-known speedy computational property.

The rest of the paper is organized as follows: in Section 2, we will be discussing the related work that has been done in this field; Section 3 is reserved for discussing the formulation process of the problem; in Section 4, we will be introducing the methodology used to carry out the analyses; the result and discussions will be presented in Sections 5 and 6; finally, Section 7 will be summarizing the conclusion that we have reached.

\section{Background}
\subsection{Related Works in Obstacle Detection}
There has been extensive research done related to building a model to assist blinded or visually impaired people; however, not a lot of work has been done on the systematic analysis of the performance of the different object detection models. These analyses of performance are important, since they present the best model for the effective building of real-world models to assist blinded and visually impaired people. In \cite{rachburee2021assistive}, the authors proposed an assistive application model for visually impaired people based on deep learning, specifically YOLOv3 with a Darknet-53 base network, installed on a smartphone. The model is trained using the Pascal VOC2007 and Pascal VOC2012 datasets and achieves high speed and accuracy in obstacle detection. The application utilizes an eSpeak synthesizer to generate audio output, allowing visually impaired individuals to interact with their surrounding environment effectively. The experimental results demonstrate the effectiveness of the proposed model in real-time obstacle detection and classification, providing safety and comfort for visually impaired individuals in their daily lives. Future work includes studying the distance between visually impaired individuals and obstacles and integrating additional theories to improve the overall application. 

The authors of \cite{said2023obstacle} mainly focused on the use of the technique of Neural Architecture Search (NAS). They proposed an intelligent navigation assistance system for visually impaired people using deep learning and NAS techniques. The deep learning model used in the system has achieved significant success through well-designed architecture. A fast NAS approach is also proposed in the paper to search for an object detection framework with efficiency in mind. The NAS is based on a tailored reinforcement learning technique. The proposed NAS is used to explore the feature pyramid network and the prediction stage for an anchor-free object detection model. The searched model was evaluated on a combination of the Coco dataset and the Indoor Object Detection and Recognition (IODR) dataset. The resulting model outperformed the original model by 2.6 percent in average precision (AP) with acceptable computation complexity. The achieved results demonstrate the efficiency of the proposed NAS for custom object detection. This motivated us to employ the YOLO-NAS model for the object detection task. 

In \cite{tapu2014real}, the researchers introduced a novel framework for detecting static/moving obstacles to assist visually impaired/blind persons in safe navigation, and the algorithm could be run in real-time on a smartphone, providing obstacle detection and classification independently. Obstacles are classified as urgent/normal based on their distance to the subject and motion vector orientation. The average processing time for obstacle detection is 18 ms/frame on an Intel Xeon Machine and 130 ms/frame on a Samsung Galaxy S4 smartphone. The paper also suggests extending the method with an object classification algorithm and converting highlighted obstacles into voice messages.

The authors of \cite{yadav2020fusion} proposed an assistive device for visually impaired people that provides automatic navigation and guidance, detects obstacles and performs real-time image processing. The device consists of a heterogeneous set of sensors and computation components, including ultrasonic sensors, a camera, a single-board DSP processor, a wet floor sensor, and a battery, and uses a machine learning model for object recognition to familiarize the user with their environment. The device can detect various obstacles such as up-stairs, down-stairs, edges, potholes, speed breakers, narrowing passages, and wet floors. The output is provided in the form of audio prompts to ensure user comfort and friendliness, and has a mean average precision (mAP) for trained objects of 81.11.

\subsection{Object Detection and other CNN-based models}
Convolutional Neural Network, also known as CNN, is a type of machine learning algorithm that is widely used in different machine learning tasks that deal with images. One of the tasks is using bounding boxes to detect objects in images. Through the learning of image data, computers are able to analyze images, recognize and classify the objects in the images into different groups. This is known as object detection. There are also several object detection algorithms, examples are the likes of R-CNN, Fast R-CNN, Faster R-Cnn, Mask R-CNN, SSD, YOLO, and more. \cite{redmon2016you,redmon2018yolov3,liu2016ssd,girshick2014rich}

R-CNN, or Region-based Convolutional Neural Network, is a computer vision algorithm that revolutionized object detection by combining the power of deep learning and region proposals. It involves extracting potential regions of interest from an image and then using a convolutional neural network to classify and localize objects within those regions. \cite{girshick2014rich}
Fast R-CNN builds upon the original R-CNN approach. It introduces a more efficient architecture by sharing convolutional features across all proposed regions, eliminating the need for redundant computations. \cite{girshick2015fast}
Faster R-CNN works by combining a Region Proposal Network (RPN) with Fast R-CNN to achieve real-time object detection with high accuracy. \cite{ren2016faster}
Another model is Mask R-CNN. The key idea behind Mask R-CNN is to generate high-quality object proposals using a region proposal network (RPN), and then refine these proposals by predicting object classes, bounding box coordinates, and pixel-level masks. \cite{he2018mask}

The other main algorithm is SSD. The key idea behind SSD is to perform object detection in a single pass of a neural network, eliminating the need for multiple stages. It achieves this by utilizing a set of predefined anchor boxes of different sizes and aspect ratios at multiple feature maps of different scales. \cite{liu2016ssd}

CNN is typically made up of convolution, pooling, and fully connected layers. The first two, convolution and pooling layers, perform the feature extraction from images, whereas the third, a fully connected layer, maps the extracted features into the final output, which are the different classes \cite{yamashita2018convolutional,gu2018recent}. There are several types of CNN architectures, including AlexNet, VGGNet, GoogLeNet, ResNet, etc.\cite{li2021survey,gu2018recent} It has various important applications in object detection, such as autonomous driving cars, facial recognition, and medical detection in healthcare \cite{vahab2019applications,masud2022smart,amit2020object}. 
% The importance of object detection rises significantly as our lives are becoming more and more convenient with the help of recognizing objects rapidly and autonomously.

\section{Problem formulation}

In this section, we mathematically formulate the obstacle detection problem. Let:

\begin{align*}
    X & : \text{Input outdoor obstacle image}, \\
    Y & : \text{Set of ground-truth annotations for objects}, \\
    y_i^{\text{class}} & : \text{Class label for object } i, \\
    y_i^{\text{box}} & : \text{Bounding box coordinates for object } i, \\
    f & : \text{Obstacle detection model \textit{i.e.}, YOLO}.
\end{align*}

The goal is to optimize the model \(f\) by minimizing the loss function:
\begin{equation*}
    \mathcal{L}(f(X), Y) = \lambda_{\text{class}} \cdot \mathcal{L}_{\text{class}}(f_{\text{class}}(X), Y) + \lambda_{\text{box}} \cdot \mathcal{L}_{\text{box}}(f_{\text{box}}(X), Y),
\end{equation*}

where:
\begin{align*}
    \mathcal{L}_{\text{class}} & : \text{Classification loss}, \\
    \mathcal{L}_{\text{box}} & : \text{Bounding box regression loss}, \\
    \lambda_{\text{class}} & : \text{Weight for classification loss}, \\
    \lambda_{\text{box}} & : \text{Weight for bounding box regression loss}.
\end{align*}

The optimization problem is to find the optimal parameters for the model \(f\) by minimizing the loss function:
\begin{equation*}
    \hat{\theta} = \arg\min_{\theta} \sum_i \mathcal{L}(f(X_i; \theta), Y_i),
\end{equation*}

where \(\theta\) represents the model parameters, and \(X_i\) and \(Y_i\) are the input outdoor image instance and ground-truth annotations for the \(i\)-th example, respectively. The optimized model $\hat{f}$ is then used to detect the obstacles in the test dataset.

\section{Methods}
\subsection{YOLO}
YOLO, with its full name being You Only Look Once, is a state-of-the-art object detection algorithm that came out back in 2016 that predicts bounding boxes and class probabilities directly from full images in one evaluation, and will predict bounding boxes for all classes in one image simultaneously, making it extremely fast \cite{mmyolo2022}.

The model divides the image into various grid cells, and it detects the objects in their center. If the center falls in one grid, then that grid is defined to be containing that object. The base YOLO model processes images in real-time at 45 frames per second, while the smaller version, Fast YOLO, achieves 155 frames per second with double the mAP of other real-time detectors. It also outperforms other detection methods, including DPM and R-CNN, when generalizing from natural images to other domains like artwork \cite{redmon2016you}.

In this paper, the implemented YOLO models included v5, v6, v7, v8, and NAS. For v5 to v8, we have only implemented 1 model from each version, and for YOLO NAS, we've implemented all three sizes of model: Small(s), Medium(m), and Large(l). Below are descriptions for each model.

\subsection{YOLOv5}

YOLOv5 \cite{jocher2022ultralytics,v5}  is a version of the YOLO (You Only Look Once) object detection model introduced in 2020 by Ultralytics. The model introduces several unique features over previous models.

First, the inclusion of TensorRT, Edge TPU, and OpenVINO enables efficient model inference across various hardware platforms. The training process is also enhanced by using retrained models that incorporate a new default one-cycle linear LR scheduler.

The support for 11 different formats extends beyond just exporting, as it facilitates inference and validation for profiling mean average precision (mAP) and speed results after the export process. Mosaic data enhancement is applied during the data input phase.

The bounding box loss function has been enhanced, transitioning from CIOU\_Loss to GIOU\_Loss, and it is employed in the prediction component.

Moreover, YOLOv5 presents a new backbone architecture called "CSPNet" (Cross-Stage Partial Network), which boosts the feature extraction procedure and enhances the accuracy of the model. Furthermore, PANet is employed in YOLOv5 to generate feature pyramids, which assist the model in efficiently managing variations in object sizes. The model head in YOLOv5 remains similar to that of the YOLOv3 and v4 versions.

\begin{figure}[t]
    \centering
    \includegraphics[width=1\linewidth]{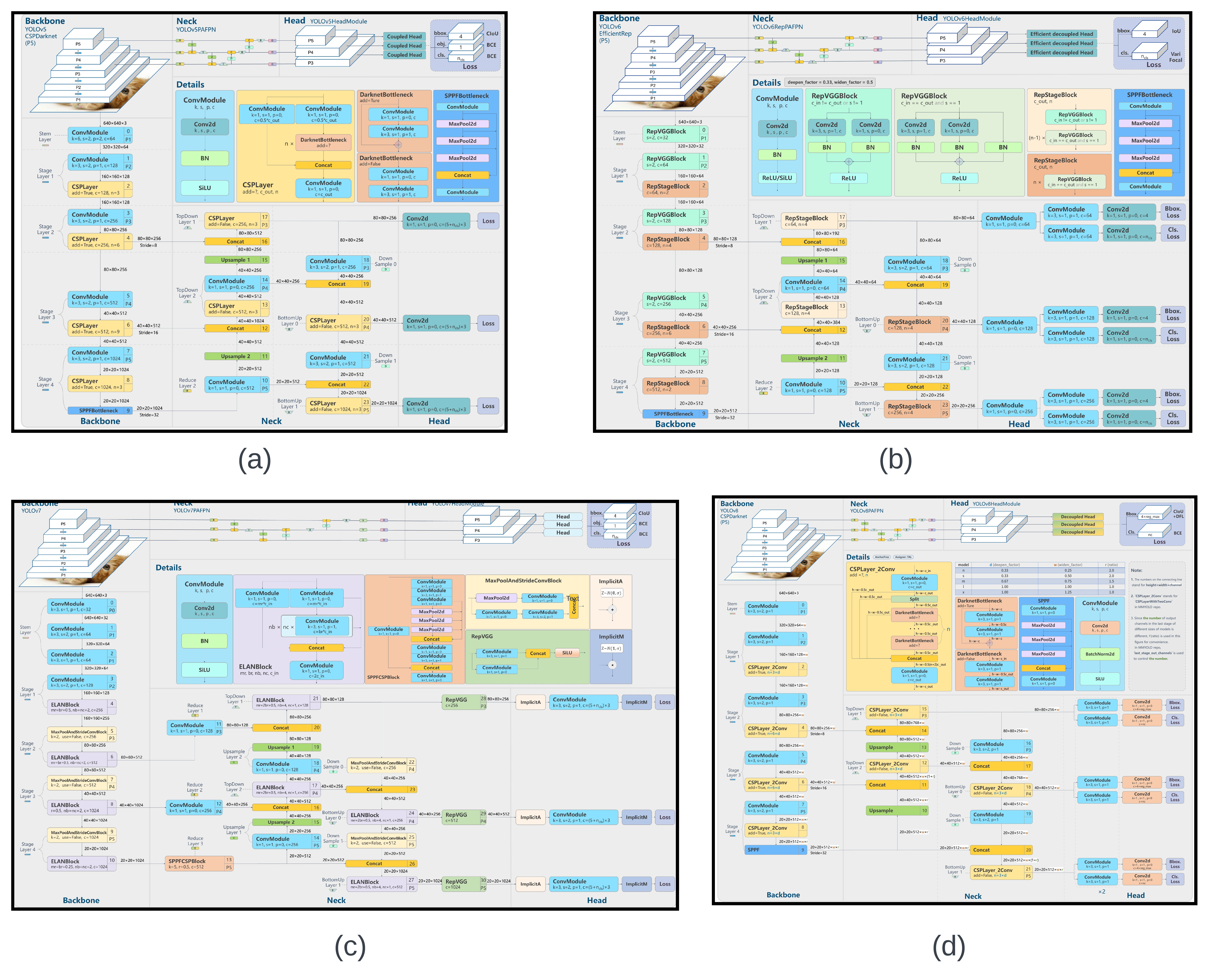}
    \caption{YOLO Architectures. (a) YOLOv5, (b) YOLOv6, (c) YOLOv7, (d) YOLOv8}
    \label{fig:enter-label}
\end{figure}

\subsection{YOLOv6}
YOLOv6 \cite{li2022yolov6,li2023yolov6}, also known as MT-YOLOv6, is a single-stage object detection model based on the YOLO architecture. It was developed by researchers at Meituan and achieves stronger performance than YOLOv5 when benchmarked against the MS COCO dataset. There are several new features that are included in this model.

First, it introduced a BiC (Bidirectional Concatenation) module in the detector's neck, which improves localization signals and provides performance improvements while maintaining minimal impact on speed.
It also introduced the Anchor-Aided Training (AAT) strategy, which combines the advantages of both anchor-based and anchor-free paradigms while maintaining efficient inference.

To improve the performance of smaller models in YOLOv6, a new self-distillation strategy is employed. This strategy enhances the auxiliary regression branch during training but removes it during inference to avoid a significant decrease in speed.

YOLOv6 also provides various pre-trained models with different scales, including quantized models for different precisions and models optimized for mobile platforms.

\subsection{YOLOv7}

YOLOv7 \cite{wang2022yolov7} is a single-stage real-time object detection model introduced in July 2022. It also hass several new features and improvements over previous versions. First, it includes a planned re-parameterization model, which is a strategy that can be applied to layers in various networks and focuses on the concept of gradient propagation path.

Also, the model introduced new techniques to enhance the training process; a new label assignment method named coarse-to-fine lead guided label assignment; and extended and compound scaling.

Additionally, YOLOv7 has been explored in the context of pose estimation. As for the performance, it achieves faster inference speeds and greater accuracy compared to its previous versions.

\subsection{YOLOv8}

YOLOv8 \cite{Jocher_YOLO_by_Ultralytics_2023,v8}, released on January 10, 2023, brought forth a range of new functionalities in comparison to its previous iterations.

First, a fresh backbone network was introduced in YOLOv8, serving as the fundamental architecture of the model.
The design facilitates a simple comparison of model performance with previous models within the YOLO family.

Then, it incorporated a novel loss function to compute the disparity between the predicted and actual values. It also implemented a novel anchor-free detection head that enables the prediction of bounding boxes without relying on predefined anchor boxes.

Peroformance-wise, YOLOv8 achieves faster inference speeds compared to other object detection models, while also maintaining a high level of accuracy. It has been used in different domains, such as wildlife detection and small object detection challenges.

\subsection{YOLO-NAS}

YOLO-NAS \cite{supergradients,Nas} represents a cutting-edge advancement in object detection, incorporating various new elements not present in earlier versions.

First of all, it introduces a novel basic block that is specifically optimized for quantization. This new block is designed to enhance the performance of quantization compared to previous versions. As a result, YOLO-NAS is able to achieve greater accuracy without sacrificing efficiency. 

It utilizes sophisticated training strategies, such as post-training quantization, AutoNac optimization, and pre-training on prominent datasets. It also takes advantage of pseudo-labeled data and gains insights from knowledge distillation by employing a pre-trained teacher model.

YOLO-NAS also demonstrated substantial enhancements in accurately detecting and localizing small objects. With a superior performance-per-compute ratio, it is well-suited for real-time edge-device applications and surpasses existing YOLO models across various datasets.

YOLO-NAS includes support for post-training quantization, which streamlines the model after the training process, resulting in improved efficiency.

It is designed to seamlessly integrate with high-performance inference engines such as NVIDIA® TensorRT™. It also enables INT8 quantization, which enhances the runtime performance to unprecedented levels.

% \begin{figure}
%     \centering
%     \includegraphics[width=1\linewidth]{YOLO-NAS.jpg}
%     \caption{YOLO-NAS Architecture\cite{supergradients}}
%     \label{fig:enter-label}
% \end{figure}

\section{Experiments and Results}
\subsection{Dataset and Preprocessing}
For this systematic empirical evaluation, we utilize the Obstacle Dataset by Wu et al. \cite{tang2023dataset} \footnote{ \url{https://github.com/TW0521/Obstacle-Dataset}.}, which included 5066 training images, 1583 testing images, and 1266 validation images. All images are photos taken from daily sidewalks, streets, and roads. The size of most images is distributed within 1500 $\times$ 1500, with a few pictures exceeding 3000 $\times$ 3000. We particularly choose this dataset as it is a comprehensive dataset also containing images from the VOC dataset, COCO dataset, and TT100K dataset. It also contains some pictures collected by the author's team in the field. As a result, the dataset can be used to validate the applicability and reliability of the models over a number of domains.
% mention image sizes and preprocessing details

There are 15 types of obstacles in this dataset, which are divided into 15 classes: Stop sign, Person, Bicycle, Bus, Truck, Car, Motorbike, Reflective cone, Ashcan, Warning column, Spherical roadblock, Pole, Dog, Tricycle, and Fire hydrant. Some sample images from the dataset are shown in Fig. 2. 

\begin{figure}
    \centering
    \includegraphics[width=1\linewidth]{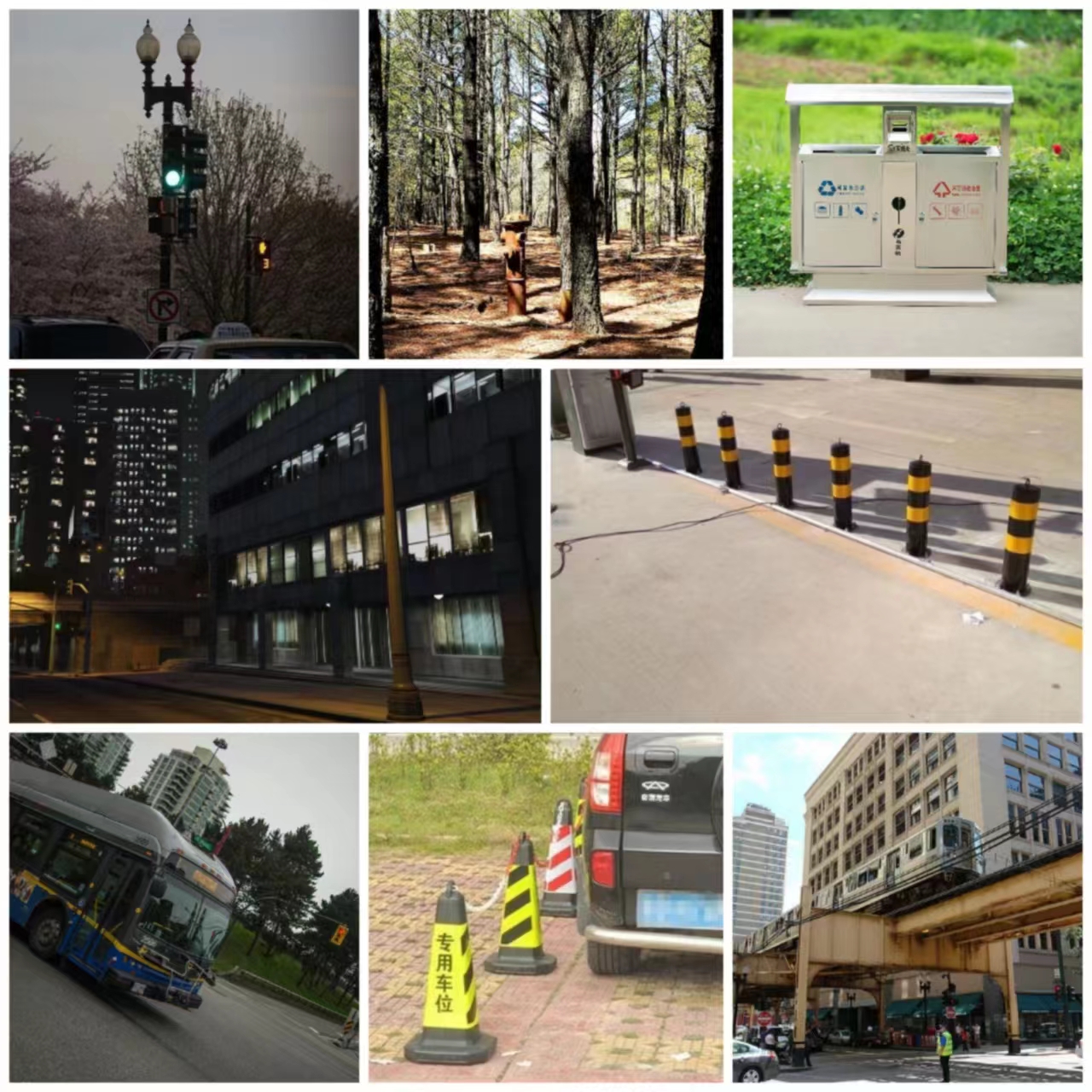}
    \caption{Sample images in the Dataset}
    \label{fig:enter-label}
\end{figure}

\subsection{Training and Implementation Details}
For all of the models, we've used a batch size of 8, and trained for 25 epochs using NVIDIA Tesla T4 GPU. We implemented and trained the models using both Google Colaboratory and Kaggle which are popular platforms for doing machine learning projects. \cite{Kaggle,colab}
% A .yaml file is also created for each of the models. Also, pre-trained weights are used from the coco dataset.

We have tweaked various hyperparameters for this work. We have set the "average best models" to True, warm-up mode as linear epoch step, initial learning rate for warm-up as 1e-6, learning rate decay factor during warmup epochs as 3, initial learning rate as 5e-4, learning rate decay mode as cosine, cosine final learning rate ratio as 0.1, optimizer as Adam, weight decay in optimizer parameters as 0.0001. We have used zero weight decay on bias and batch-normalization and leveraged exponential moving averaging  with decay factor as 0.9 and decay type as threshold with the "mixed precision" set to True.

\subsection{Performance Metrics}
The Metrics that we have used in our research are Confusion Matrix, Precision, Recall, and Mean Average Precision (mAP). Additionally, we use F1 score to further evaluate YOLO-NAS models and analyse reasons behind its suboptimal performance.

% In a binary classification, there are two classes that machines can classify the images into, which can be named to be a positive class, and a negative class, or class 0 and class 1. For those predictions that are predicted correctly out of all of the objects classified to be the Positive class, they are referred to True Positive; for those predictions that are predicted correctly out of all of the objects classified to be the Negative class, they are referred to True Negative. Both of these are correct predictions. Model errors occur when the model makes a wrong prediction, and these are False Positive and False Negative. False Positive happens when the model makes a wrong prediction in the predicted positive class, or class 0, and False Negative happens when the model makes a wrong prediction in the predicted negative class, or class 1.

% Using a confusion matrix, we put all four of these in a single table, and visualize the relationships in between, as shown in Fig. 7.

% \begin{figure}
%     \centering
%     \includegraphics[width=1\linewidth]{confusion_matrix.png}
%     \caption{Illustration of Confusion Matrix \cite{confusion_matrix}}
%     \label{}
% \end{figure}

Precision measures the percent of correct predictions out of the predicted Positive class, and the formula for precision is expressed in Eqn. 1. 

\begin{equation}
    Precision = \frac{TP}{TP + FP} 
\end{equation}
where $TP$ and $FP$ denote True Positive and False Positive respectively.

Recall measures the percent of correct predictions out of all of those real instances. The formula is expressed in Eqn. 2: 

\begin{equation}
    Recall = \frac{TP}{TP + FN} 
\end{equation}
where $FN$ denotes a False Negative.

However, these are not limited to binary classification; they can also be used in muti-class classification as shown in Eqn. 3 and 4.

\begin{equation}
    Precision\ in\ Multi-class = \frac{TP\ in\ all\ class}{TP + FP\ in\ all\ classes} 
\end{equation}

\begin{equation}
    Recall\ in\ Multi-class = \frac{TP\ in\ all\ classes}{TP + FN\ in\ all\ classes} 
\end{equation}

The Mean Average Precision (mAP) is also a metric to analyze the performance of the model, and to calculate this, we'll need to first graph the precision vs. recall curve, and then get the area under that curve. This will be the average precision for one class. Thus, averaging out the area under the curve for all classes will get the mAP.

For YOLO-NAS, we also use F1 score. Often, precision and recall offer a trade-off, \textit{i.e.}, one comes at the cost of the other. Thus, F1 score combines the harmonic mean of both of these values to get a more accurate performance evaluation, simultaneously maximizing both the precision and recall. The formula for calculating F1 score is shown in Eqn. 5.
\begin{equation}
    F1 = 2 * \frac{Precision * Recall}{Precision + Recall} 
\end{equation}

All these relationships can be visualized using a confusion matrix.
\subsection{Performance Analysis}
In this section, we report the performance achieved by all the YOLO models.

\subsubsection{Results for YOLOv5}
YOLOv5 reached an overall precision of 78.1$\%$, recall of 68.2$\%$, and mAP@0.5 of 74.2$\%$. Table 1 shows the classwise performance of YOLOv5.
The class with the highest precision was Reflective Cone, achieving a precision of 90.4$\%$, whereas the Spherical Roadblock was the class with the highest Recall and mAP@0.5, reaching 91.8$\%$ and 93$\%$ respectively. The class with the lowest precision and mAP is Truck, at 58.5$\%$ and 51.1$\%$ respectively; Pole is the class with the lowest recall, only at 41.1$\%$.

\begin{table}[]
    \centering
    \caption{Classwise performance analysis for YOLOv5}
    \begin{tabular}{ c | c | c | c | c | c }
    \toprule
     Class & Images & Labels & Precision & Recall & mAP@.5 \\ 
     \midrule
      All & 1262 & 7938 & 0.781 & 0.682 & 0.742 \\ 
      Stop Sign & 1262 & 216 & 0.793 & 0.782 & 0.834 \\
      Person & 1262 & 2205 & 0.76 & 0.581 & 0.653\\
      Bicycle & 1262 & 263 & 0.798 & 0.525 & 0.61\\
      Bus & 1262 & 199 & 0.77 & 0.588 & 0.683 \\
      Truck & 1262 & 428 & 0.585 & 0.481 & 0.511\\
      Car & 1262 & 1990 & 0.716 & 0.594 & 0.659\\
      Motorbike & 1262 & 223 & 0.609 & 0.579 & 0.622\\
      Reflective Cone & 1262 & 462 & 0.904 & 0.82 & 0.877 \\
      Ashcan & 1262 & 283 & 0.875 & 0.815 & 0.871 \\
      Warning Column & 1262 & 315 & 0.683 & 0.832 & 0.828\\
      Spherical Roadblock & 1262 & 280 & 0.898 & 0.918 & 0.93\\
      Pole & 1262 & 660 & 0.717 & 0.411 & 0.558 \\
      Dog & 1262 & 135 & 0.816 & 0.696 & 0.748 \\
      Tricycle & 1262 & 154 & 0.895 & 0.832 & 0.911\\
      Fire Hydrant & 1262 & 125 & 0.898 & 0.776 & 0.832\\
     \bottomrule
    \end{tabular}
    \label{tab:my_label}
\end{table}

\subsubsection{Results for YOLOv6}
YOLOv6 achieved an average precision of 59$\%$ for IoU@0.5:0.95 and for Area of "All". The average recall for IoU@0.5-0.95 and Area of "All" is 71.7$\%$. More details are provided in Table 2.

\begin{table}[]
    \centering
    \caption{Performance analysis for YOLOv6}
    \begin{tabular}{ c | c | c | c }
    \toprule
     Metrics & IoU & Bounding Box Area & Scores \\ 
     \midrule
      Average Precision & 0.5:0.95 & All & 0.59 \\ 
      Average Precision & 0.5 & All & 0.784 \\
      Average Precision & 0.75 & All & 0.651\\
      Average Precision & 0.5:0.95 & Small & 0.239\\
      Average Precision & 0.5:0.95 & Medium & 0.506\\
      Average Precision & 0.5:0.95 & Large & 0.721\\
      Average Recall & 0.5:0.95 & All & 0.717\\
      Average Recall & 0.5:0.95 & Small & 0.465\\
      Average Recall & 0.5:0.95 & Medium & 0.675\\
      Average Recall & 0.5:0.95 & Large & 0.818\\
     \bottomrule
    \end{tabular}
    \label{tab:my_label}
\end{table}

\subsubsection{Results for YOLOv7}

\begin{table}[]
    \centering
    \caption{Classwise performance analysis for YOLOv7}
    \begin{tabular}{ c | c | c | c | c | c }
    \toprule
     Class & Images & Labels & Precision & Recall & mAP@.5 \\ 
     \midrule
      All & 1262 & 7938 & 0.786 & 0.778 & 0.817 \\ 
      Stop Sign & 1262 & 216 & 0.844 & 0.819 & 0.874 \\
      Person & 1262 & 2205 & 0.787 & 0.687 & 0.749\\
      Bicycle & 1262 & 263 & 0.756 & 0.672 & 0.728\\
      Bus & 1262 & 199 & 0.793 & 0.754 & 0.8\\
      Truck & 1262 & 428 & 0.549 & 0.629 & 0.626\\
      Car & 1262 & 1990 & 0.765 & 0.709 & 0.764\\
      Motorbike & 1262 & 223 & 0.653 & 0.704 & 0.727\\
      Reflective Cone & 1262 & 462 & 0.905 & 0.892 & 0.944 \\
      Ashcan & 1262 & 283 & 0.915 & 0.835 & 0.904 \\
      Warning Column & 1262 & 315 & 0.75 & 0.898 & 0.898\\
      Spherical Roadblock & 1262 & 280 & 0.902 & 0.952 & 0.958\\
      Pole & 1262 & 660 & 0.625 & 0.636 & 0.642 \\
      Dog & 1262 & 135 & 0.772 & 0.748 & 0.822 \\
      Tricycle & 1262 & 154 & 0.877 & 0.928 & 0.947\\
      Fire Hydrant & 1262 & 125 & 0.893 & 0.802 & 0.871\\
     \bottomrule
    \end{tabular}
    \label{tab:my_label}
\end{table}
Overall, for all of the classes, YOLOv7 achieved a precision of 78.6$\%$, a recall of 77.8$\%$, and a mAP@0.5 of 81.7$\%$.
Table 3 demonstrates the classwise performance of YOLOv7.
The class with the highest Precision is class "Ashcan", reaching 91.5$\%$. The class with the highest Recall and mAP@0.5 is also Spherical Roadblock, reaching 95.2$\%$ for recall and 95.8$\%$ for mAP@0.5. The class with the worst performance is Truck, where it only achieved a precision of 54.9$\%$, a recall of 62.9$\%$, and a mAP@0.5 of 62.6$\%$.

\subsubsection{Results for YOLOv8}
\begin{table}[]
    \centering
     \caption{Classwise performance for YOLOv8}
    \begin{tabular}{ c | c | c | c | c | c | c}
    \toprule
    Classes & Images & Instances & Precision & Recall & mAP@.5 & mAP@.5-.95 \\ 
     \midrule
     All & 1262 & 7938 & 0.8 & 0.683 & 0.758 & 0.561 \\ 
     Stop sign & 1262 & 216 & 0.82 & 0.773 & 0.832 & 0.698 \\
     Person & 1262 & 2205 & 0.797 & 0.574 & 0.677 & 0.417 \\
     Bicycle & 1262 & 263 & 0.791 & 0.532 & 0.624 & 0.389 \\
     Bus & 1262 & 199 & 0.738 & 0.558 & 0.659 & 0.521 \\
     Truck & 1262 & 428 & 0.683 & 0.473 & 0.542 & 0.369\\
     Car & 1262 & 1990 & 0.799 & 0.599 & 0.7 & 0.478 \\
     Motorbike & 1262 & 223 & 0.671 & 0.605 & 0.685 & 0.43 \\
     Reflective Cone & 1262 & 462 & 0.92 & 0.818 & 0.896 & 0.715 \\
     Ashcan & 1262 & 283 & 0.907 & 0.792 & 0.87 & 0.746 \\
     Warning Column & 1262 & 315 & 0.856 & 0.797 & 0.872 & 0.621 \\
     Spherical Roadblock & 1262 & 280 & 0.913 & 0.9292 & 0.953 & 0.789 \\
     Pole & 1262 & 660 & 0.636 & 0.597 & 0.625 & 0.339 \\
     Dog & 1262 & 135 & 0.773 & 0.593 & 0.717 & 0.499 \\
     Tricycle & 1262 & 154 & 0.882 & 0.89 & 0.916 & 0.763 \\
     Fire Hydrant & 1262 & 125 & 0.809 & 0.745 & 0.807 & 0.636 \\
     \bottomrule
    \end{tabular}
   
    \label{tab:my_label}
\end{table}
The experimental results for YOLOv8 model are demonstrated in Figures 8, 9, 10, and Table 4.

\begin{figure}
    \centering
    \includegraphics[width=1\linewidth]{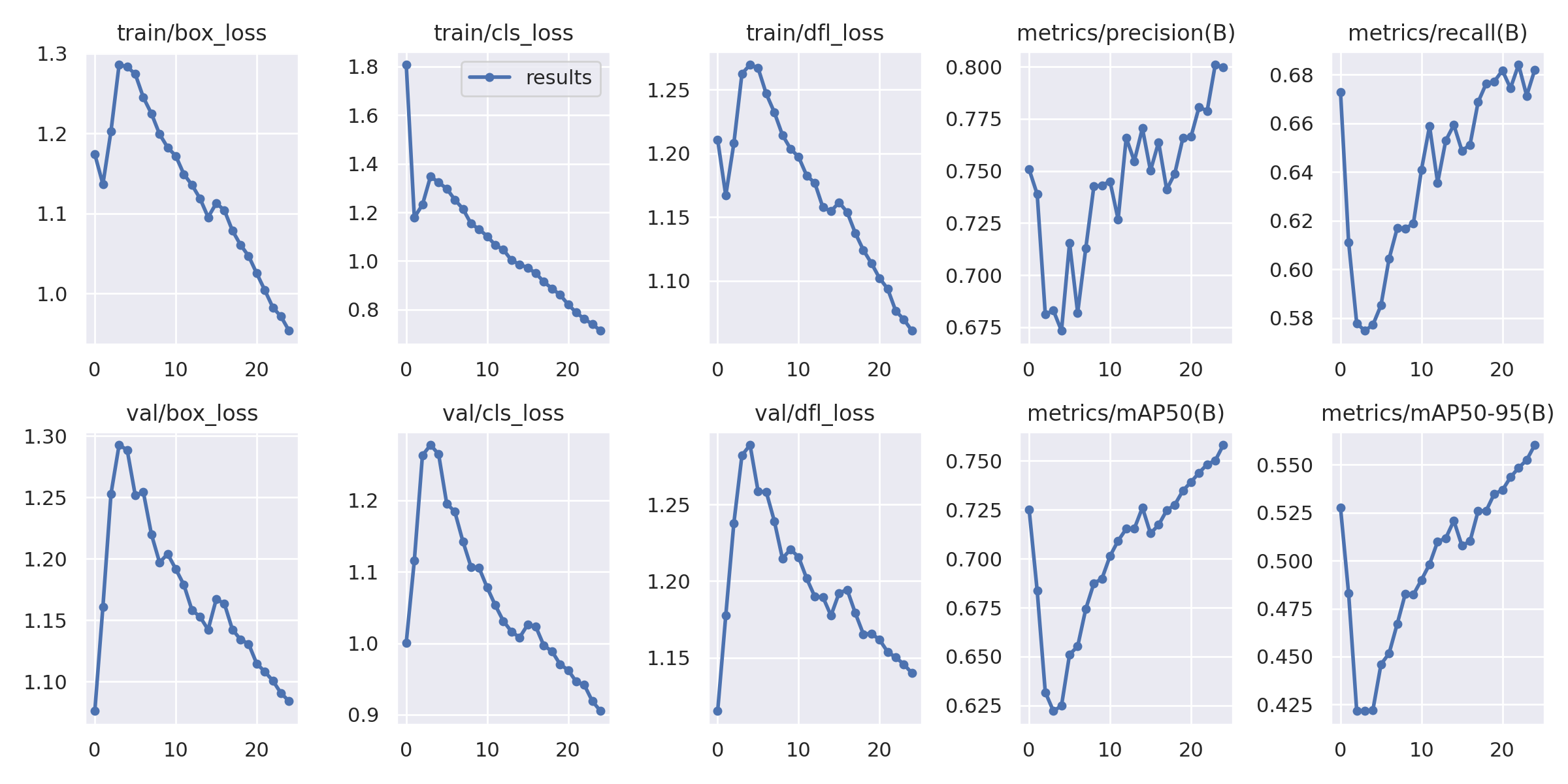}
    \caption{Loss and performance curves for YOLOv8: The first and second rows correspond to the training and validation curves respectively. The first three columns show the individual losses - box loss (which represents how well the algorithm can cover over the real object), cls loss (Classification loss, which tells how well the model can predict the correct class of a given object in a bounding box), and dfl loss (or Objectness loss, which checks the probability that an object exists in a proposed region of interest) \cite{kasper2021detecting}, whereas the last two columns demonstrate two metrics - precision and recall}
    \label{fig:enter-label}
\end{figure} 

\begin{figure}
    \centering
    \includegraphics[width=1\linewidth]{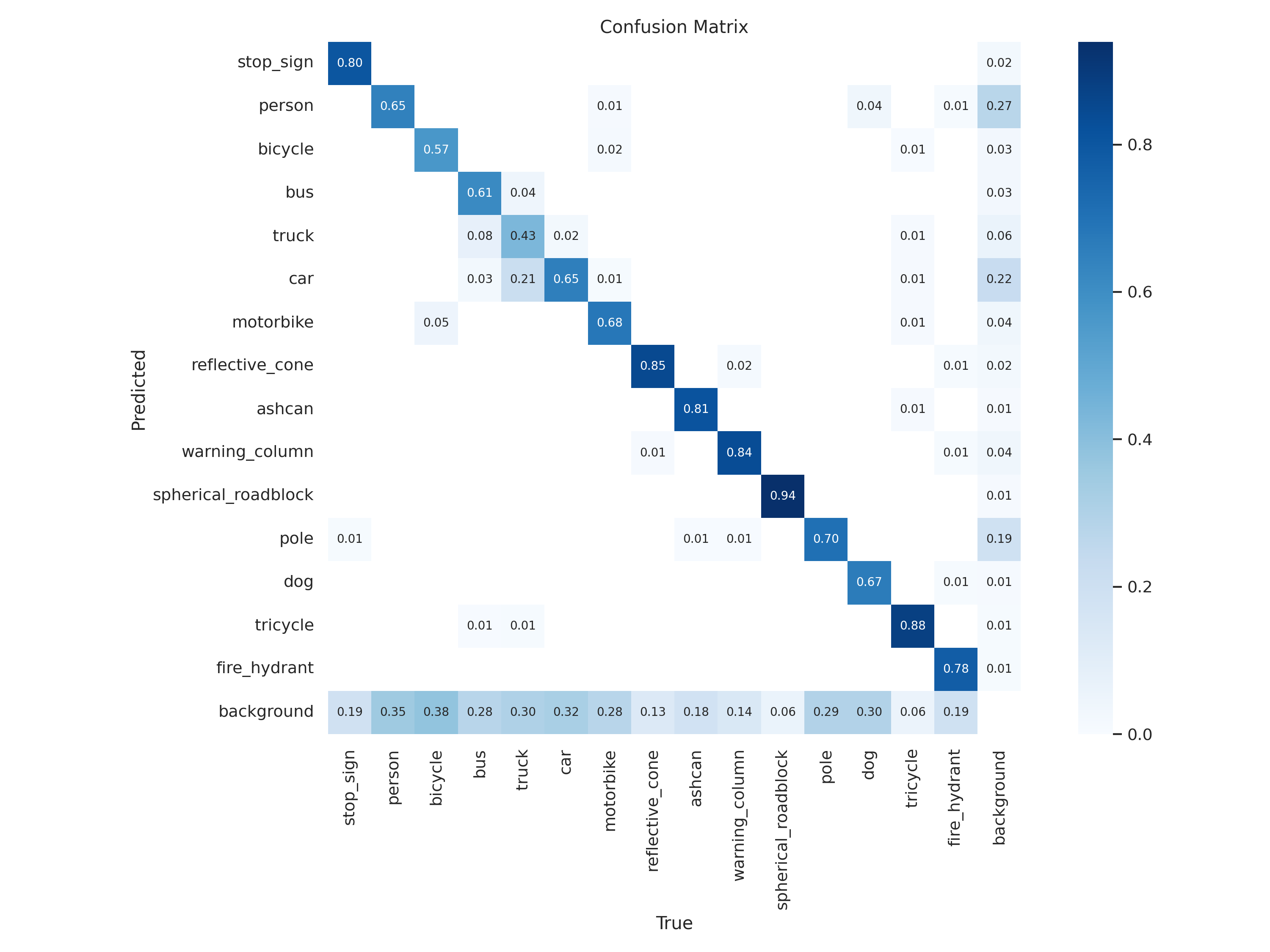}
    \caption{Confusion Matrix depicting the performance of YOLOv8 for the obstacle detection task with 15 classes. As it is clearly evident, the highest true positive value is recorded as 0.94 for the class 'spherical roadbloack' whereas the lowest true positive is obeserved to be 0.43 for the class 'truck'.}
    \label{fig:enter-label}
\end{figure}

\begin{figure}
    \centering
    \includegraphics[width=1\linewidth]{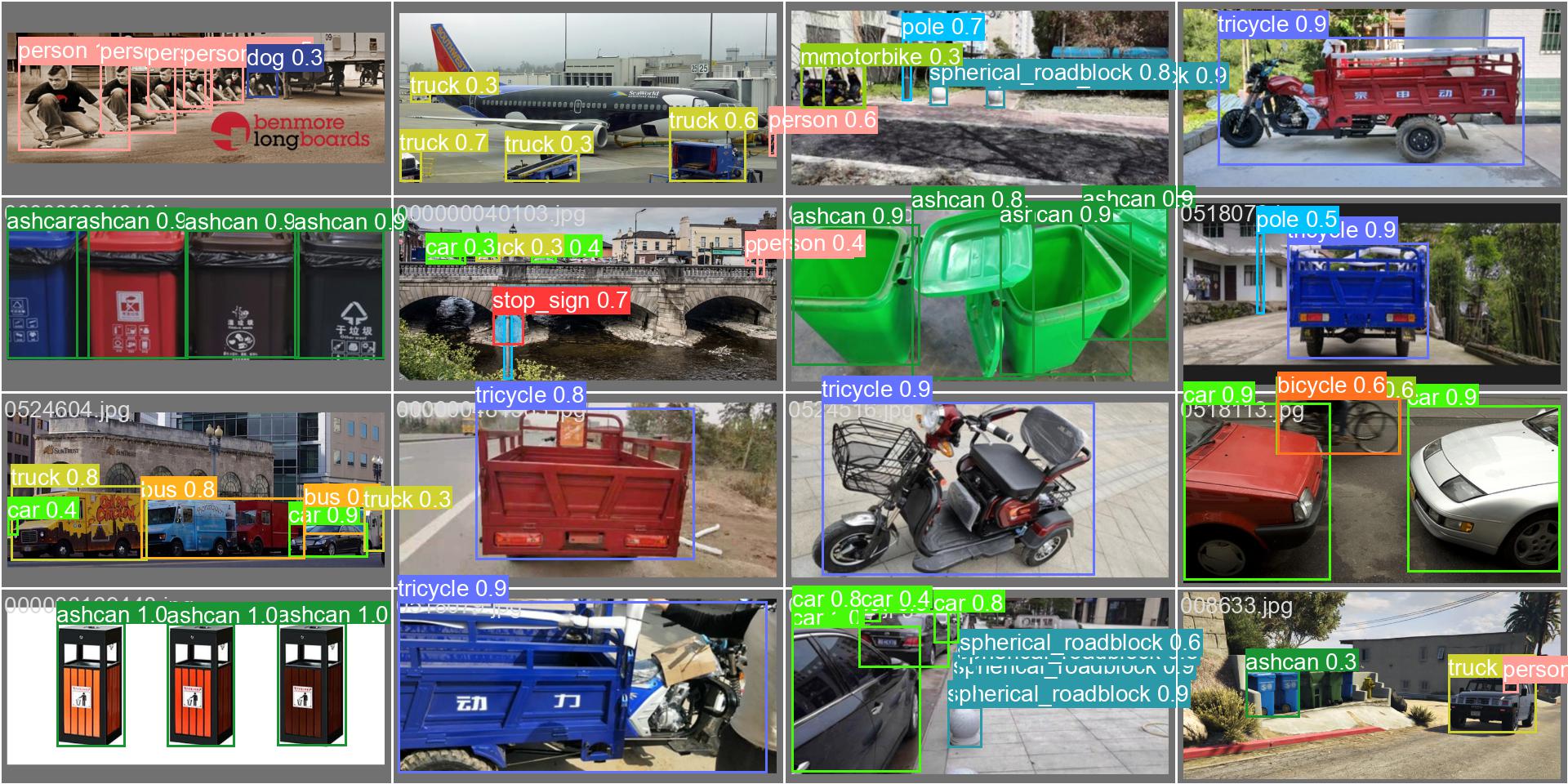}
    \caption{ Visualization of the performance of YOLOv8 with bounding boxes and corresponding obstacle classes predicted by the model along with the respective confidence scores}
    \label{fig:enter-label}
\end{figure}

Figure 9 shows the Confusion Matrix for YOLOv8. For this Confusion Matrix, the numbers that are on the diagonal line from top left to bottom right are the ones that are correctly predicted. The rows represent the predicted classes made by the machine, while the column represents the ground truth classes.

The overall results for all classes for YOLOv8 are: a precision at 80$\%$, a recall at 68.3$\%$, a mAP@0.5 at 75.8$\%$, and a mAP@0.5-0.95 at 56.1$\%$.
The class with the best performance is Speherical Roadblock, achieving a Precision of 91.3\%, a Recall of 92.9$\%$, a mAP@0.5 of 95.3$\%$, and a mAP@0.5-0.95 of 78.9$\%$. The class with the lowest precision is Pole, where it only reaches 63.6$\%$; the lowest recall and mAP@0.5 class is Truck, with a recall of 47.3$\%$ and a mAP@0.5 of 54.2$\%$. The lowest class with mAP@0.5-0.95 is class Pole, achieving only 33.9$\%$.

\subsubsection{Results for YOLO-NAS}

YOLO-NAS has three models: Large, Medium, and Small. The model with the highest precision NAS-S, achieving a precision of 78.8\%. The model with the highest recall, mAP, and F1 is NAS-M, achieving a recall of 62.7\%, a mAP@0.5 of 69.8\%, and an F1@0.5 of 67.85\%. Full results, including the losses calculated from the different loss functions, can be found in table 5.

\begin{table}[]
    \centering
        \caption{Analysis of Performance for YOLO-NAS}
    \begin{tabular}{ c | c | c | c | c | c | c | c | c }
    \toprule
    Models & Loss\_cls & Loss\_iou & Loss\_dfl & Loss & Precision@0.5 & Recall@0.5 & mAP@0.5 & F1@0.5 \\ 
     \midrule
     NAS-Large & 0.8801 & 0.1687 & 0.8863 & 1.753 & 0.7853 & 0.6274 & 0.698 & 0.6785 \\
     NAS-Medium & 0.8843 & 0.1746 & 0.8546 & 1.7726 & 0.7589 & 0.6148 & 0.6829 & 0.6723 \\
     NAS-Small & 0.8947 & 0.1766 & 0.8936 & 1.7896 & 0.7888 & 0.5941 & 0.6673 & 0.6523 \\
     \bottomrule
    \end{tabular}
    \label{tab:my_label}
\end{table}

\subsubsection{Overall Results}
The performance of each of the models is demonstrated in Table 6. 
\begin{table}[]
    \centering
    \caption{Comparative Performance Analysis for different YOLO Models}
    \begin{tabular}{ c | c | c | c }
    \toprule
    Models/Result & Precision@0.5 & Recall@0.5 & mAP@0.5 \\ 
     \midrule
     YOLOv5 & 0.781 & 0.682 & 0.742 \\
     YOLOv6 & 0.786 & 0.714 & 0.784 \\
     YOLOv7 & 0.786 & 0.778 & 0.817 \\
     YOLOv8 & 0.80 & 0.682 & 0.758 \\
     YOLO-NAS-Small & 0.7888 & 0.5941 & 0.6673 \\
     YOLO-NAS-Medium & 0.7875 & 0.6206 & 0.6906 \\
     YOLO-NAS-Large & 0.7853 & 0.6274 & 0.698 \\
     \bottomrule
    \end{tabular}
    \label{tab:my_label}
\end{table}
The model with the highest precision is YOLOv8, achieving a precision of 80\%; the model with the highest recall and mAP is YOLOv7, reaching 77.8\% of recall and 81.7\% mAP respectively. The F1 score is only available for YOLO-NAS, and the highest is NAS-L, at 67.85\%. The model with the lowest precision is YOLOv5, at 78.1\%; the lowest recall, mAP and F1 is YOLO-NAS-S, achieving a recall of 59.41\%, a mAP of 66.73\%, and a F1 of 65.23\%.

\subsection{Ablation Studies}
Through comprehensive experiments via grid search on different hyperparameters, we found the YOLO-NAS model to be highly sensitive to the threshold score. Therefore, we tweaked the model by varying the threshold score to analyze its performance. We found that this parameter has a positive relationship with precision, while having a negative relationship with recall. As the threshold went higher, the precision increased, while the recall decreased and vice versa. Table 7 shows the results of the ablation analysis using three threshold values: 0.3, 0.5, and 0.7. At a threshold value of 0.7, the YOLO-NAS Medium model produces the highest precision of 93.16$\%$, while the Small model produces the lowest recall of 42.46$\%$. At a threshold value of 0.3, the Large model produces the highest recall, at 78.48$\%$, and the Medium model produces a precision of 44.87$\%$. This shows that there is a trade-off between Precision and Recall and hence the threshold value needs to be carefully designed based on the given dataset and task.

\begin{table}[]
    \centering
    \caption{Results for Ablation Studies by varying threshold score (denoted as Score\_thres) for small, medium, and large versions of the YOLO-NAS Model}
    \begin{tabular}{ c | c | c | c | c | c }
    \toprule
    Model & Score\_thres value & Precision@0.5 & Recall@0.5 & mAP@0.5 & F1@0.5 \\ 
     \midrule
     Large & 0.7 & 0.9247 & 0.4564 & 0.702 & 0.5679 \\ 
     Medium & 0.7 & 0.9316 & 0.4387 & 0.6934 & 0.555 \\
     Small & 0.7 & 0.929 & 0.4246 & 0.6673 & 0.5357 \\
     Large & 0.5 & 0.7853 & 0.6274 & 0.698 & 0.6785 \\ 
     Medium & 0.5 & 0.7875 & 0.6206 & 0.6906 & 0.6723 \\
     Small & 0.5 & 0.7888 & 0.5941 & 0.6673 & 0.6523 \\
     Large & 0.3 & 0.4938 & 0.7848 & 0.7017 & 0.5898 \\ 
     Medium & 0.3 & 0.4487 & 0.7718 & 0.6818 & 0.5543 \\
     Small & 0.3 & 0.5088 & 0.7585 & 0.6725 & 0.5636 \\
     \bottomrule
    \end{tabular}
    \label{tab:my_label}
\end{table}

\section{Discussion and Conclusion}
In this paper, we investigated seven YOLO models for the task of detecting outdoor obstacles on sidewalks. We found the model with the highest precision is YOLOv8, with a precision reaching up to around 80$\%$. The model with the highest recall and mAP is YOLOv7, reaching 77.8$\%$ and 81.7$\%$ respectively. The YOLO model with the lowest precision is YOLOv5, but still, it reached a precision of 78.1$\%$. The lowest recall and mAP is shown by YOLO-NAS-S, with a recall of 59.41$\%$ and an mAP of 66.73$\%$. We also performed some ablation studies on the YOLO-NAS models, revealing the trade-off of precision and recall based in the threshold score. When threshold score increases (from 0.5 to 0.7), the precision increases, while the recall decreases and vice-versa. Thus, the threshold score has a positive relationship with precision, while having a negative relationship with recall and needs to be carefully tuned. Our code will be made publicly available upon acceptance.

There are several areas that could be modified and investigated further as the extension of this study. First, even after fine-tuning, tweaking and ablation analysis, the performance of YOLO-NAS is not optimal. Future research could delve deeper into it and adapt it to obstacle detection applications. 

Second, now that the best performing YOLO model is known, the work can be extended to include other object detection models and compared with the best YOLO models to investigate comparative performance of different object detection algorithmic families on outdoor obstacle detection in sidewalks. 
Finally, real-time object detection devices could be built using YOLOv8 to help visually impaired people navigate through the real world.

\bibliography{sn-bibliography}% common bib file

\end{document}